# Incorporating Causal Prior Knowledge as Path-Constraints in Bayesian Networks and Maximal Ancestral Graphs

**Giorgos Borboudakis**  BORBUDAK@ICS.FORTH.GR
**Ioannis Tsamardinos**  TSAMARD@ICS.FORTH.GR
Institute of Computer Science, FORTH, Greece, and Computer Science Department, University of Crete, Greece

## Abstract

We consider the incorporation of causal knowledge about the presence or absence of (possibly indirect) causal relations into a causal model. Such causal relations correspond to directed paths in a causal model. This type of knowledge naturally arises from experimental data, among others. Specifically, we consider the formalisms of Causal Bayesian Networks and Maximal Ancestral Graphs and their Markov equivalence classes: Partially Directed Acyclic Graphs and Partially Oriented Ancestral Graphs. We introduce sound and complete procedures which are able to incorporate causal prior knowledge in such models. In simulated experiments, we show that often considering even a few causal facts leads to a significant number of new inferences. In a case study, we also show how to use real experimental data to infer causal knowledge and incorporate it into a real biological causal network. The code is available at mensxmachina.org.

## 1. Introduction

Qualitative causal knowledge, such as $X$ causally affects $Y$ (denoted as $X \dashrightarrow Y$) or $X$ does not causally affect $Y$ (denoted as $X \not\dashrightarrow Y$) is often available in many domains. It may stem from expert or domain knowledge (the methylation levels of a gene's promoter $X$ causally reduces its expression $Y$) or known semantic or temporal constraints (e.g., demographic variables do not causally affect gender). Such knowledge may also come from small-sample experiments where a quantity is manipulated: if temperature $X$ is var-ied in a yeast culture, then all (non-)differentially expressed genes are (not) causally affected by temperature. These relations can be identified by simple hypotheses tests, even if one cannot robustly induce a complete causal model due to a small sample size.

In this paper, we devise theory and algorithms for incorporating a given set **K** of $X \dashrightarrow Y$ and $X \not\dashrightarrow Y$ relations into a causal model. As causal models we consider Bayesian Networks (BNs) and Maximal Ancestral Graphs (MAGs) and their respective Markov equivalence classes Partially Directed Acyclic Graphs (PDAGs) and Partially Oriented Ancestral Graphs (PAGs). Such models can be induced from data by learning algorithms such as the PC and the FCI (Spirtes et al., 2000). MAGs are a generalization of BNs that admit possible latent confounders. Typically, when learning from observational data, several statistically indistinguishable models are consistent with the data forming a Markov Equivalence class. These models share the same edges but may disagree on their orientations. In these models causal facts of the form $X \dashrightarrow Y$ and $X \not\dashrightarrow Y$ correspond to the presence and absence of a directed path, respectively.

First, we characterize the Markov equivalence class of all BNs (MAGs) that belong in the given PDAG (PAG) and at the same time are consistent with **K**. It turns out that this type of equivalence class cannot be represented with a PDAG (PAG) but a simple extension of these formalisms is required that we name *Path-Constrained PAG (PDAG)* (PC-PAG, PC-PDAG). A PC-PDAG (PC-PAG) is similar to a PDAG (PAG) with the addition of new types of edges denoting the presence or absence of a directed path. In general, the incorporation of **K** into a PDAG (PAG) forces the orientation of certain edges and results in a corresponding PC-PDAG (PC-PAG) with fewer structural uncertainties. As a simple example consider that given the PAG $X \circ\!\!-\!\!\circ Y \circ\!\!-\!\!\circ Z$ and knowledge $\mathbf{K} = \{X \dashrightarrow Z\}$ one can infer the PC-PAG $X \to Y \to Z$ (which also happens to be a PAG and a MAG in this case).

Subsequently, we develop algorithms that given a PDAG (PAG) $\mathcal{P}$ and a set of knowledge facts $\mathbf{K}$ discover all implied edge orientations and return the corresponding PC-PDAG (PC-PAG), if $\mathcal{P}$ and $\mathbf{K}$ are consistent. We show that the algorithms are computationally more efficient than brute force algorithms that enumerate all BNs (MAGs) in the equivalence class of $\mathcal{P}$ to identify the ones that are also consistent with $\mathbf{K}$. Later on, we extend the algorithms to deal with cases where $\mathcal{P}$ and $\mathbf{K}$ are inconsistent.

In simulated experiments with randomly generated networks as well as real networks appearing in the literature, we show that often, even for small $|\mathbf{K}|$, a large number of orientations is made possible. This provides evidence for the utility of identifying and using this type of prior knowledge in causal discovery. We also present a case study where we incorporate causal knowledge induced from real biological data to a known biological network.

Several other methods that address prior knowledge for causal discovery have appeared in the literature. These methods can incorporate knowledge on the parameters of the network (Niculescu et al., 2006), on the presence or absence of *direct* relations (Meek, 1995), on a total ordering of the variables (Cooper & Herskovits, 1992), or the complete structure of the network (Heckerman et al., 1995). *Direct* causal relations in a model (i.e., not mediated by any other modeled variable) correspond to edge in the model; being "direct" depends on the context (i.e., the modeled variables). In contrast, path-constraints do not depend on the context and are semantically different. In the yeast example of the first paragraph, one may deduce that temperature is causally affecting a gene expression, but not necessarily directly: other genes may mediate the effect. In (O'Donnell R. T., 2008) a method is presented for incorporating possibly indirect relations, but relies on computationally expensive Markov Chain - Monte Carlo (MCMC) simulations. No prior algorithm (see (Borboudakis et al., 2011) for an early effort) can incorporate causal knowledge of possibly indirect relations for MAGs.

## 2. Background

We briefly review some background preliminaries, assuming the reader's familiarity with causal modeling. Maximal Ancestral Graphs (MAGs) are graphical models that represent causal relations among a set of variables, as well as probabilistic properties, such as conditional independencies. A key property of MAGs is that they are able to model latent confounders and selection variables without explicitly introducing them into the model, using bi-directed and un-directed edges respectively. For this paper, we do not consider cases of selection variables (i.e., we actually consider what is called Directed Maximal Ancestral Graphs (DMAGs) that do not have un-directed edges) (Spirtes et al., 2000; Richardson & Spirtes, 2002). We will refer to DMAGs as MAGs for ease of notation.

MAGs contain two kinds of edges: directed edges ($\rightarrow$) and bi-directed edges ($\leftrightarrow$). Each edge has two marks (or orientations), tails (-) and/or arrowheads ($>$). A wildcard mark ($*$) can be a tail or an arrowhead. Edge $A* \rightarrow B$ is *into* $B$, and edge $A \rightarrow B$ is *out of* $A$.

A *path* in a MAG $M$ is a sequence of distinct vertices $\langle V_0, V_1, \ldots, V_n \rangle$, s.t. for $0 \leq i < n$, $V_i$ and $V_{i+1}$ are adjacent in $M$. A path is *directed* if for $0 \leq i < n$, $V_i \rightarrow V_{i+1}$ is present in $M$. $A$ is an *ancestor* of $B$ and $B$ a *descendant* of $A$ if $A = B$ or there is a directed path from $A$ to $B$ in $M$. A *directed cycle* occurs in $M$ if $B \rightarrow A$ and $A$ is an ancestor of $B$. An *almost directed cycle* occurs in $M$ if $B \leftrightarrow A$ and $A$ is an ancestor of $B$. A triple $\langle X, Y, Z \rangle$ is said to form a *collider* if $X$ and $Z$ are into $Y$.

MAGs, by definition, do not contain any directed or almost directed cycles. As a consequence, an arrowhead denotes non-ancestry, whereas a tail denotes ancestry. Specifically, a directed edge $A \rightarrow B$ denotes that $A$ is a causal parent of $B$, whereas a bi-directed edge $A \leftrightarrow B$ denotes that neither of the two variables is a causal ancestor of each other; in addition, when faithfulness holds (defined below) the bi-directed edge denotes $A$ and $B$ share a latent common cause (confounder). Next a graphical criterion called *m-separation* is defined, which connects the graph with properties of the joint distribution of the data.

**Definition 2.1** (m-separation). *In a MAG, a path $p$ between vertices $A$ and $B$ is m-connecting relative to (condition to) a set of vertices $\mathbf{Z}$, $(A, B \notin \mathbf{Z})$ if: (i) every non-collider on $p$ is not a member of $\mathbf{Z}$, (ii) every collider on $p$ is an ancestor of some member of $\mathbf{Z}$. $A$ and $B$ are said to be m-separated by $\mathbf{Z}$ if there is no m-connecting path between them relative to $\mathbf{Z}$.*

We assume that the *Markov Condition* and the *Faithfulness Condition* hold for MAGs, i.e., $A$ and $B$ are m-separated by $\mathbf{Z}$ if and only if $A$ and $B$ are independent given $\mathbf{Z}$. So, one can graphically determine which independencies hold in the data distribution. In addition, it is required by definition of MAGs that for every missing edge between $A$ and $B$ there exists a subset of the variables $\mathbf{Z}$ s.t. $A$ and $B$ are m-separated by $\mathbf{Z}$.

It may be the case that two or more different MAGs share the same $m$-separations. Those MAGs are said to be *Markov equivalent*.

**Definition 2.2** (Markov Equivalence). *Two MAGs $M_1, M_2$, with the same set of vertices, are Markov equivalent if for all disjoint sets of vertices $\mathbf{A}, \mathbf{B}, \mathbf{Z}$, where $\mathbf{A}$ and $\mathbf{B}$ are not empty, $\mathbf{A}$ and $\mathbf{B}$ are m-separated by $\mathbf{Z}$ in $M_1$ if and only if they are m-separated by $\mathbf{Z}$ in $M_2$.*

All Markov equivalent MAGs form a Markov equivalence class. A Partially Oriented Ancestral Graph (PAG) represents such a Markov equivalence class. PAGs contain three kinds of marks: arrowheads (>), tails (-) and circles (◦). It has the same adjacencies as any member of the equivalence class, and every non-circle mark is invariant in any member of the equivalence class. Arrowheads and tails have the same semantics as in MAGs. Circles denote *uncertainties*; both orientations appear in some MAGs of the equivalence class. A path is *possibly directed* if there is an orientation of the uncertainties of the PAG which creates a directed path. A triple $\langle X, Y, Z\rangle$ forms a *definite non-collider* if $X$ and $Z$ are not adjacent and $X, Z$ are not both into $Y$. FCI (Spirtes et al., 2000) is an asymptotically correct algorithm for learning a PAG.

Bayesian Networks (BNs) are special cases of MAGs with no bi-directed edges; thus, latent causes of two or more modeled variables (confounders) cannot be represented so that Faithfulness holds in a way that is also consistent with the causal semantics of the edges. The Markov equivalence class of BNs is called Partially Directed Acyclic Graphs (PDAG; some authors use the term essential graph instead and PDAG for a different type of graph). In the rest of the paper, we develop the theory and algorithms for the general case (PAGs) and discuss specializations for PDAGs.

## 3. Problem Definition

We assume that a PAG $\mathcal{P}$ defined over variables $\mathbf{V}$ is given representing a Markov equivalence class of MAGs faithful to some distribution over $\mathbf{V}$. $\mathcal{P}$ may contain structural uncertainties about the direction of some edge-points. $\mathcal{P}$ could be induced from data by a learning algorithm such as FCI, or be otherwise known and fixed. In addition, we are given a set of prior knowledge constraints $\mathbf{K}$ of the form $X \dashrightarrow Y$ (we call these *positive constraints*) or $X \not\dashrightarrow Y$ (*negative constraints*), where $X, Y \in \mathbf{V}$. Thus, *knowledge constraints must concern modeled variables*. A constraint $X \dashrightarrow Y$ ($X \not\dashrightarrow Y$) implies that $X$ is (is not) a causal ancestor of $Y$, i.e., there must (must not) be a directed path $X \rightarrow \cdots \rightarrow Y$ in $\mathcal{P}$. Conversely, given that the network is faithful to some distribution, if a path is present (absent) then $X$ is causing (not causing) $Y$.

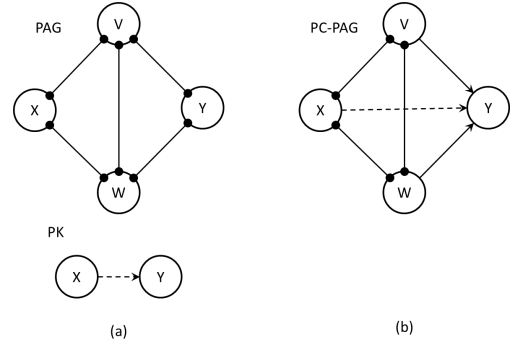

*Figure 1.* (a) Input PAG and prior knowledge constraint. (b) Corresponding PC-PAG. The orientations $V \rightarrow Y \leftarrow W$ are inferred; edge $X \dashrightarrow W$ is necessary to exclude MAGs that are not consistent with prior knowledge.

Thus, *each piece of knowledge in $\mathbf{K}$ corresponds to a path constraint about the presence or absence of a directed path in $\mathcal{P}$*. Finally, we assume that $\mathcal{P}$ and $\mathbf{K}$ are consistent, i.e., the path-constraints induced by the latter can be satisfied by at least one MAG in $\mathcal{P}$ (later we develop an algorithm removing this assumption).

As an example, assume that we are given PAG $X \circ\!\!-\!\!\circ Y \circ\!\!-\!\!\circ Z$. Incorporating knowledge $\mathbf{K} = \{X \dashrightarrow Z\}$ one can infer that $X \rightarrow Y \rightarrow Z$. Instead, if $\mathbf{K} = \{X \not\dashrightarrow Z\}$ then one can infer $X \leftarrow \circ Y \circ\!\!-\!\!\circ Z$.

Notice that in this setting knowledge is qualitative (the strength of the causal effect is not represented) and both $\mathcal{P}$ and $\mathbf{K}$ are assumed correct (their uncertainty is not represented). Since $\mathcal{P}$ is assumed correct and it encodes all conditional dependencies and independencies in the joint distribution: (a) there is no point representing and including in $\mathbf{K}$ knowledge about the data dependencies and independencies; they are either already represented in $\mathcal{P}$ or are inconsistent with it. (b) neither positive nor negative constraints add to our knowledge about the independencies in $\mathcal{P}$ and *thus they cannot affect the skeleton of $\mathcal{P}$, and only reduce our uncertainty about edge marks*.

We now consider how to represent the set of MAGs that are both Markov equivalent with PAG $\mathcal{P}$ and satisfy prior knowledge $\mathbf{K}$. We call this set *Path-Constrained PAG*: $\mathbf{PC\text{-}PAG}(\mathcal{P}, \mathbf{K})$. It is easy to check that a PC-PAG is an equivalence class. As we show next, a PC-PAG cannot be represented by a PAG, thus *we need to define a new graphical object to represent the class*.

A simple example is now described. Consider the PAG $\mathcal{P}$ and $\mathbf{K} = \{X \dashrightarrow Y\}$ in Figure 1(a). Any MAG in PC-PAG($\mathcal{P}, \mathbf{K}$) must have the orientations $V \rightarrow Y \leftarrow W$: if $Y \ast\!\!\rightarrow V$, then $V \rightarrow X$ because $\langle X, V, Y \rangle$ is a definite non-collider. Since $X \dashrightarrow Y$ the only remaining option is to orient $X \to W \to Y$, in which case either a cycle or an almost directed cycle is created. Thus, $Y \ast \to V$ does not hold, i.e., $V \to Y$. Symmetrically, we get $W \to Y$. Figure 1(b) (ignoring the dash edge momentarily) shows a PAG $\mathcal{P}'$ consistent with $\mathcal{P}$ and the new orientations. $\mathcal{P}'$ admits 19 possible MAGs, out of which only 9 are also consistent with **K**. It is impossible to further orient any edge in a way that the PAG admits these 9 MAGs and only those. Thus, *PAGs are not closed under the addition of path-constraints*. We now present a type of graph that can represent a PC-PAG equivalence class; as for PAGs we overload the term PC-PAG to indicate both the class and its graphical representation. The new type of graph contains additional edges to represent the path constraints.

**Definition 3.1.** *A* Path-Constrained PAG *of PAG $\mathcal{P}$ and knowledge **K**, PC-PAG($\mathcal{P}$, **K**) is a graph with two types of edges: solid and dashed edges, s.t., (i) the skeleton of the solid edges is the same as $\mathcal{P}$, (ii) each solid edge-mark is either tail (-) or arrowhead (>) if the corresponding feature is invariant in all MAGs in $\mathcal{P}$ also consistent with **K**, and $\circ$ otherwise, and (iii) an indirect edge $X \dashrightarrow Y$ ($Y \circ \dashrightarrow X$) is present if $X \dashrightarrow Y$ ($X \not\dashrightarrow Y$) is in **K** but there is not a directed path (there is a possible directed path) from $X$ to $Y$ using solid edges only.*

*The vertices and the solid edges form a PAG that we call* the underlying PAG. *We similarly define* Path-Constrained PDAG *and its corresponding underlying PDAG, when $\mathcal{P}$ is a PDAG instead of a PAG.*

Figure 1(b) is the PC-PAG of the PAG and knowledge in Figure 1(a). The independencies shared by all member MAGs in a PC-PAG can be read off the underlying PAG. The dashed edges also denote ancestral relations so the graph remains ancestral. There can be at most one edge between a pair of vertices. For a given $\mathcal{P}$ and **K** the corresponding PC-PAG is unique. We now define the following problem:

**Problem 1.** *Given a PAG $\mathcal{P}$ over variables **V** and a set of causal prior knowledge $\mathbf{K} = \{\mathcal{K}\}_{i=1}^{M}$, where each $K_i$ is of the form $A \dashrightarrow B$ or $A \not\dashrightarrow B$, $A, B \in \mathbf{V}$, induce the PC-PAG($\mathcal{P}$, **K**).*

## 4. Algorithms for Consistent Prior Knowledge

We now develop algorithms that identify the PC-PAG or PC-PDAG given a baseline $\mathcal{P}$ and knowledge **K**. We will be referring to the general case of PC-PAGs, unless we need to explicitly differentiate. By definition, the skeleton of the solid edges of PC-PAG is the same as the one in $\mathcal{P}$; once the edge marks of the solid edges are determined, the indirect edges are trivially determined by **K**. Thus, the main objective should be to determine the edge marks of the solid edges, i.e., the orientations shared by all MAGs in $\mathcal{P}$ also consistent with **K**.

---

**Algorithm 1** Find-PC-PAG($P$,$K$)

1: **Input:** *PAG $P$; set of causal prior knowledge $K$*
2: **Output:** *boolean* sat; *PC-PAG C*
3:
4: global Found
5: **for each** un-oriented mark $X \ast - \circ Y$ in $P$ **do**
6:     Found$(X, Y, >) = false$
7:     Found$(X, Y, -) = false$
8: **end for**
9: $sat = \text{Search}(P, K)$
10: **if** $\neg sat$ **then return** $\langle sat, \emptyset \rangle$ **end if**
11: $C = P$
12: **for each** un-oriented mark $X \ast - \circ Y$ in $P$ **do**
13:     **if** Found$(X, Y, >) \wedge \neg$ Found$(X, Y, -)$ **then**
14:         Orient$(C, X, Y, >)$
15:     **else if** $\neg$ Found$(X, Y, >) \wedge$ Found$(X, Y, -)$ **then**
16:         Orient$(C, X, Y, -)$
17:     **end if**
18: **end for**
19: **for each** non-satisfied $K_i \in \mathbf{K}$ in $P$ **do**
20:     **if** $K_i$ is of type $X \dashrightarrow Y$ **then**
21:         Add edge $X \dashrightarrow Y$ to $C$
22:     **else**
23:         Add edge $X \dashleftarrow \circ Y$ to $C$
24:     **end if**
25: **end for**
26: **return** $\langle sat, C \rangle$

---

**Algorithm 2** Search($P$,$K$)

1: **Input:** *PAG $P$; set of causal prior knowledge $K$*
2: **Output:** *boolean* sat
3:
4: **if** $\neg Valid(P, K)$ **then return** *False* **end if**
5: **if** PruneRule($P$) **then return** *True* **end if**
6: $\langle X, Y \rangle = $ any $X \ast \!-\!\!\circ Y$ in $P$
7: **if** there is no such edge $\langle X, Y \rangle$ **then**
8:     UpdateFound($P$)
9:     **return** *True*
10: **end if**
11: $P_1 = $ ApplyOrientation($P, X, Y, >$)
12: $sat_1 = Search(P_1, K)$
13: $P_2 = $ ApplyOrientation($P, X, Y, -$)
14: $sat_2 = Search(P_2, K)$
15: **return** $sat_1 \vee sat_2$

Algorithm 1 starts from a given PAG $\mathcal{P}$ and keeps adding orientations until it is converted to a MAG. It does so recursively so as to explicitly or implicitly enumerate all consistent MAGs and identify the invariant edge marks. The data structure Found$(X, Y, m)$ stores a flag indicating whether a MAG has been found where the right end-point of edge $X$—$Y$ is marked as $m$. The procedure **Search** performs the actual search and computes the values of the identified edge-marks in Found. If all MAGs identified agree on a given edge-mark $m$ of edge $X$—$Y$, $m$ is transferred to the output graph $\mathcal{C}$ by calling procedure Orient$(\mathcal{C}, X, Y, m)$. Once the edge-marks of $\mathcal{C}$ are determined the algorithm inserts the dashed edges by applying the PC-PAG definition. The algorithm is sound and complete provided the search procedure identifies all edge-marks that belong in at least one consistent MAG.

We now focus on the search procedure. The search strategy is essential for the computational efficiency of the algorithm. **We actually present 4 different search procedures, that are all encoded in the same pseudo-code of Algorithm 2** due to space limitations. The algorithms perform search with and without pruning. In addition, sub-procedures of Algorithm 2 may be specialized for PDAGs or PAGs. Hence, there are 4 different specialization of the pseudo-code for each of the above combinations. Algorithm 2 accepts parameters $\mathcal{P}$ and **K** and returns the corresponding PC-PAG, or a flag indicating no consistent MAG was found.

**Search-No-Pruning**. This version of Algorithm 2 does not include line 5. For each ∘ mark in an edge $X*$—∘$Y$, the edge may be oriented as $X \leftarrow Y$ or $X*\rightarrow Y$ in a MAG of the class. The procedure performs a *chronological backtracking search with forward checking* (Dechter, 2003), i.e., it recursively calls itself for each possible way to place an edge mark (Lines 12 and 14), while propagating these decisions to eliminate inconsistent choices. Thus, in the worst case the algorithm calls itself at most $2^{\#u}$, where $\#u$ is the number of uncertainties (∘ marks in the input graph $\mathcal{P}$). The procedure stops in two cases:

(i) an orientation has been made that leads to an invalid MAG. Procedure **Valid**$(\mathcal{P}', \mathbf{K})$ determines validity by checking the following three conditions:

(1) There are no directed cycles nor almost directed cycles (the latter is only checked for PC-PAGs).

(2) No prior knowledge constraint is violated.

(3) The set of $m$-separations justifying each missing edge in $\mathcal{P}$ remains the same in $\mathcal{P}'$. This set of $m$-separations can be stored by FCI or similar algorithms as $\mathcal{P}$ is induced; otherwise it can be found from $\mathcal{P}$ in a preprocessing step. The reason one needs to check this condition is because orientations imposed during search may change the set of discriminating paths (Spirtes et al., 2000), and thus the independence model of $\mathcal{P}'$ may be different (in our implementation this conditions is only checked if there are no more uncertainties in the graph). This condition is checked only for PC-PAGs and not for PC-PDAGs (as we will see later in the ApplyOrientation procedure).

(ii) there are no more uncertainties in the graph and we have found a MAG $\mathcal{P}'$ (line 7); if it is valid, then procedure **UpdateFound**$(\mathcal{P}')$ sets Found$(X, Y, -) = True$ or Found$(X, Y, >) = True$, if the edge $X \leftarrow Y$ or $X*\rightarrow Y$ is present in $\mathcal{P}'$, respectively.

**Forward Checking**. During search, an orientation of an edge mark may imply other orientations. For example, if vertices $A, B, C$ form a definite non-collider triple and $A$ is oriented into $B$, the edge between $B$ and $C$ has to be oriented out of $B$ and into $C$. This implicitly prunes the search tree, similar to unit propagation in SAT solving algorithms. Procedure **ApplyOrientation**$(\mathcal{P}, X, Y, m)$ applies mark $m$ (either $-$ or $>$) to edge $X*$—∘$Y$ and propagates the orientation. For the case of PC-PDAGs one can simply apply Meek's rules (Meek, 1995) until convergence *to find all implied orientations*. Unfortunately, for the case of PC-PAGs there is no known complete procedure (Zhang notes this as an open problem (Zhang & Spirtes, 2005), p. 81). In this case, we use rules $R1$-$R3$ of FCI (Zhang & Spirtes, 2005) to do some, but not all of possible propagations. As a result it is possible to generate a MAG that does not belong to the Markov equivalence class represented by the initial PAG. Thus, Condition 3 in procedure **Valid** above is necessary to check. Application of the propagation rules takes polynomial time.

**Search-with-Pruning**. We now present a condition that allows early stopping of the search without sacrificing completeness and significantly improves the efficiency. For a call of Search$(\mathcal{P}', \mathbf{K})$, let us call with $A$(ssigned) the set of assigned orientations so far in the search path, i.e., $A = \{\langle X, Y, m\rangle$ s.t. the end-point at Y is oriented$\}$; let $U$(nassigned) be the set of orientations remaining, i.e., $U = \{\langle X, Y\rangle$ s.t. $X*$—∘$Y \in \mathcal{P}'\}$. Then note that,

**Rule 1** (Prune Rule). *If for each mark in $A$, a MAG has already been found, and for each unassigned mark in $U$ a MAG has been found for all possible orientations, there is no need to proceed with search and the procedure returns* True. *No matter what the orientations in $U$ we end up with, our knowledge of possible*

Incorporating Causal Prior Knowledge as Path-Constraints in Causal Models

*orientations will not increase. This check is performed by procedure* **PruneRule**.

When pruning, selecting to recurse on the marks for which a consistent MAG has not been found yet may lead to earlier pruning. Thus, as a heuristic in line 6, we give preference to edges with such marks. In section 6 we present results showing that pruning leads to an exponential speed up of the algorithm.

## 5. Dealing with Inconsistent Prior Knowledge

Algorithm 1 returns *False* if the given PAG $\mathcal{P}$ and prior knowledge **K** are inconsistent. Ideally in this case, one should express the uncertainty in both and infer new orientations in a probabilistic yet efficient way. Such a procedure however, is still eluding us. As an approximation, we now present Algorithm 3 that identifies a subset $\mathbf{K}' \subseteq \mathbf{K}$ that is consistent with $\mathcal{P}$ and maximizes a score function denoting preferences on the prior knowledge.

For each piece of knowledge $K_i \in \mathbf{K}$ we denote with $u_i$ the utility of satisfying it in a MAG, and $c_i$ the cost (penalty) of not satisfying it. We can then define the score function of satisfying knowledge $\mathbf{K}' \subseteq \mathbf{K}$:

$$Sc(\mathbf{K}', \mathbf{K}) = \sum_{K_i \in \mathbf{K}'} u_i + \sum_{K_i \in \mathbf{K} - \mathbf{K}'} c_i$$

By setting all $u_i$ to 1 and $c_i$ to 0, the algorithm will find the largest subset of consistent prior knowledge constraints. With the given setup one can also handle cases where each prior knowledge constraint has a prior belief $p_i$ assigned to it. Specifically, if $u_i = \log(p_i)$ and $c_i = \log(1-p_i)$ and one assumes these probabilities are independent, then the $Sc(\mathbf{K}', \mathbf{K})$ corresponds to the prior probability $P(\mathbf{K}')$. In general of course, these probabilities will not be independent since $P(X \dashrightarrow Y \dashrightarrow Z) = 1 \implies P(X \dashrightarrow Z) = 1$.

Algorithm 3 is a branch-and-bound algorithm that does not branch if the current search path cannot possibly lead to a MAG with higher score than what has already been found. Given a PAG $P'$ in the current search node, let $\mathbf{K}^s$ be the set of currently satisfied prior knowledge constraints, $\mathbf{K}^v$ the set of violated constraints and $\mathbf{K}^r$ the set of all remaining constraints in **K**. An upper bound on the best score to find under this search path can then be computed as follows:

$$ScBound(\mathbf{K}^r, \mathbf{K}^s, \mathbf{K}^v) = \sum_{K_i \in \mathbf{K}^s} u_i + \sum_{K_i \in \mathbf{K}^v} c_i + \sum_{K_i \in \mathbf{K}^r} \max(u_i, c_i) \quad (1)$$

**Algorithm 3** SearchBnB($P$,**K**)

  **Input:** *PAG $P$; set of causal prior knowledge* **K**
  **Output:** *score maximizer* $\mathbf{K}' \subseteq \mathbf{K}$ *stored globally*

  global $S'$, maximum score found, initialized to $-\infty$
  global $\mathbf{K}'$, score maximizer, initialized to $\emptyset$

  **if** $\neg$Valid($P,K$)$\vee$MaxPosScore($P,\mathbf{K}$) $\leq S'$ **then**
     **return**
  **end if**
  $\langle X, Y \rangle =$ any $X \ast\!\!-\!\!\circ Y$ in $P$
  **if** there is no such edge $\langle X, Y \rangle$ **then**
     $S' = $ Score($P, \mathbf{K}$)
     $K' = $ FindSatisfied($P, \mathbf{K}$)
     **return**
  **end if**
  $P_1 =$ApplyOrientation($P, X, Y, >$)
  SearchBnB($P1, \mathbf{K}$)
  $P_2 =$ApplyOrientation($P, X, Y, -$)
  SearchBnB($P_2, \mathbf{K}$)

Procedures **Score** and **MaxPosScore** compute score $Sc$ and upper bound $ScBound$ given the PAG $P$ in the current search node and knowledge **K**. Once a leaf of the search has been reached, i.e., all orientations are determined and we have reached a MAG $P'$ in the current node, procedure **FindSatisfied** is called to identify the subset of **K** that is satisfied in $P'$.

## 6. Experimental Evaluation

**Speed-Up of Pruning Rule**. We evaluated the performance gain of our method when the pruning is on. We randomly generated Bayesian Networks for varying numbers of vertices and edges. For PAGs 20% of the vertices were randomly picked to be hidden. The networks were then converted to their corresponding PAG or PDAG. The numbers of vertices were {5,10,15} for PAGs and {50,100,150} for PDAGs. Smaller networks were generated for PAGs than PDAGs because: (a) PAGs usually have many more uncertainties than PDAGs for the same size and settings of the generation process and (b), the algorithm for PAGs is slower in general than for PDAGs given the same number of uncertainties, because orientations lead to more propagations in PDAGs than PAGs. For PAGs we set an upper limit of 50 on the number of uncertainties to avoid computationally intractable problems. The edge density varied between 10-90%, with steps of 1 (a total of 81 different densities). The number of prior knowledge constraints were {1,2,3,5,7,10} for PAGs and {1,2,3,5,7,10,15,20} for PDAGs. The constraints were sampled from the set of unknown and consistent pairwise causal relations of the given model (i.e., they were not already satisfied). Figure 2 shows the mean number of invocations of the algorithm (search nodes) vs the number of uncertainties of each PAG. The y-axis is logarithmically scaled. The *effective branching factor* $b$ when pruning is on is smaller than when it is not used, leading to exponential computational savings. We also compared our algorithm with a one-sample t-test (using the difference of invocations of the methods), and obtained p-values of $9 \cdot 10^{-3}$ and $8 \cdot 10^{-16}$ for the PDAG and PAG methods respectively, showing that using pruning offers statistically significant improvements.

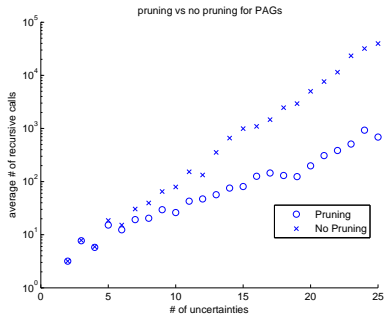

Figure 2. Comparing Algorithm 2 with and without pruning for randomly generated PAGs and knowledge constraints. Pruning leads to exponential computational savings: the effective branching factor for PAGs is approximately 1.25 vs. 1.5 when pruning is off, and for PDAGs is approximately 1.1 vs 1.2 when pruning is off.

**Evaluation of Inference Capabilities**. We evaluated the ability of our methods to infer new orientations. We randomly generated networks as before. The number of vertices was {10,15,20,25,30} and {50,100,150,200,300,500}, whereas the number of prior knowledge facts was {1,2,3,5,7,10,15,20,25} and {1,2,3,5,7,10,15,20,30,50} for PAGs and PDAGs respectively. We also ran experiments for 3 real networks which are commonly used in the literature (Alarm, Hailfinder and Child). For the real networks, the number of prior knowledge constraints were within {1,2,3,5,7,10,15,20,30,50}, both for PDAGs and PAGs. The selection of prior knowledge was repeated 100 times. That is a total of $3 \cdot 10 \cdot 100 = 3000$ runs, for each type of model (PDAGs and PAGs). We measure the ability to make novel inferences, called *Inference Rate*, as $IR = \frac{\#inferences}{\#uncertainties}$ where #inferences is the number of mark orientations inferred by incorporating the knowledge and #uncertainties the number of ◦ marks in the input PAG or PDAG. Figures 3(a) and 3(b) show the mean inference rate for PAGs and PDAGs respectively, as the number of prior knowledge constraints increases. In general, (a) inference rate is significant (more than 30%) even for a small number of constraints (e.g., 10) and (b) inference rate is higher in PDAGs than PAGs everything else being equal.

**A Case-Study with Real Data**. We obtained two flow-cytometry datasets of (K. Sachs, 2005). Both datasets measure a set of 11 protein concentrations on the same biological pathway under different experimental conditions. The first dataset contains 707 samples and the (indirectly) manipulated variable is PKA, whereas the second contains 913 samples and the (indirectly) manipulated variable is PKC. Causal prior knowledge is inferred as follows: for manipulated protein $M$, we infer the constraint $M \dashrightarrow X$ for every $X$ with a p-value of a Spearman correlation less than 0.01, and $M \not\dashrightarrow X$, when the p-value is greater than 0.5 This leads to 11 causal constraints as knowledge **K**. We also consider a portion of the biological pathway as our gold standard MAG (Figure 4(a)) that we convert to its corresponding PAG $\mathcal{P}$ (Figure 4(b)). **K** and $\mathcal{P}$ are inconsistent in this case because two constraints in **K** are due to statistical errors. To select a consistent set of prior knowledge, we ran Algorithm 3 with weights $u_i = \log(1 - p_i)$ and $c_i = \log(p_i)$, where $p_i$ is the p-value of a positive constraint, or 1-p-value for a negative constraint. The algorithm selected a consistent subset $\mathbf{K}'$ of size 6 that was then incorporated into $\mathcal{P}$ via Algorithm 1. $\mathbf{K}'$ still included one statistical error. The resulting PC-PAG is shown in Figure 4(c). It orients 11 out of 22 initial edge mark uncertainties, out of which one is erroneous due to the false negative constraint in $\mathbf{K}'$ with a p-value of 0.86.

## 7. Conclusion

We present algorithms for incorporating path-constraints to PDAGs and PAGs corresponding to known, possibly indirect, causal relations. The algorithms use chronological backtracking search, with forward checking and a pruning rule stemming from the semantics of the graphs. A branch-and-bound variation of the algorithms for dealing with knowledge inconsistent with the given PDAG or PAG is also presented. Our experimental results show that typically even a few causal constraints can orient a significant number of edges. In a case study we show how experimental studies (where some variables are manipulated) can be used to infer such causal constraints and be incorporated into an incomplete (i.e. with structural uncertainties) causal model. The algorithms could form a basis for extensions that take into consideration degrees of belief on each constrain or network feature.

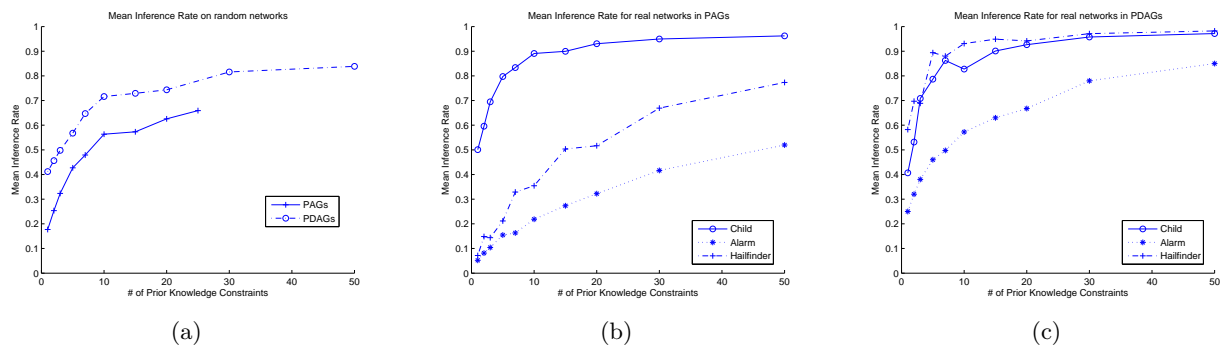

*Figure 3.* (a) Inference rates in randomly generated graphs. The rate increases more sharply in PDAGs than in PAGs: the absence of confounding variables leads to more propagations of the knowledge constraints. In both cases, even a small number of knowledge constraints (e.g., 10) leads to a significant percent of new inferences (more than 30% of the edge uncertainties. (b and c) The inference rate in real networks highly depends on the network, particularly for PAGs; the rate is again much higher for PDAGs than PAGs.

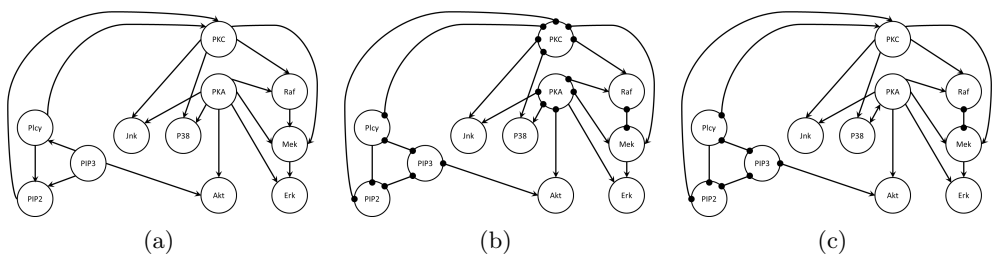

*Figure 4.* (a) A part of a biological signal pathway taken from (K. Sachs, 2005) considered as the gold standard in the case-study. The variables denote protein concentration levels. (b) The PAG of the network. (c) Causal knowledge **K** was inferred from two of the experimental flow cytometry datasets in (K. Sachs, 2005). **K** was inconsistent with the network. A consistent subset **K**′ of size 6 was selected via Algorithm 3 and was incorporated into the network by Algorithm 1. Out of the initial 22 uncertain orientations the algorithm inferred the orientation of 11 of them.

## Acknowledgements

We thank the reviewers for their helpful comments. This work was partially funded by the REACTION GA 248590 EU project.